# Pattern Recognition and Memory Mapping using Mirroring Neural Networks

Name(s) Dasika Ratna Deepthi [(1)] and K. Eswaran [(2)]

*Address: (1) Osmania University, Hyderabad - 500 007*
*(2) Sreenidhi Institute of Science and Technology, Yamnampet, Ghatkesar, Hyderabad - 501 30, India*

*Email addresses radeep07@gmail.com, kumar.e@gmail.com*

## Abstract

*In this paper, we present a new kind of learning implementation to recognize the patterns using the concept of Mirroring Neural Network (MNN) which can extract information from distinct sensory input patterns and perform pattern recognition tasks. It is also capable of being used as an advanced associative memory wherein image data is associated with voice inputs in an unsupervised manner. Since the architecture is hierarchical and modular it has the potential of being used to devise learning engines of ever increasing complexity.*

## 1: INTRODUCTION

In this paper, we introduce an algorithm using Mirroring Neural Networks (MNN) which performs a dimension reduction of input data followed by mapping, to recognize patterns. There have been many investigations done on pattern recognition, a few of which deal with geometric feature extraction [1], manifold learning [2] and non-linear dimensional reduction [3] and [4] etc.,. In addition to pattern recognition through data reduction, the neural network approach can also be used to resolve high dimensional problems in clustering [5] and to study complex neuronal properties [6].

We in our approach, develop an architecture which does non-linear data reduction associated with mapping using a special type of neural network called MNN (refer [7] and [8] for details). The architecture is hierarchical; possesses the ability of unsupervised learning and in its form is an imitation of the basic natural neural system [9]. We follow the bottom-up approach that resembles the architecture used by Dileep George and Jeff Hawkins [10] and [11] to recognize the patterns by organizing the MNN modules (approximating cortical regions of the cortical system) of different levels using connections [12]. Our conviction is that the proposed architecture can be used as a part of an unsupervised hierarchical learning machine to be used for complex real-world pattern recognition problem, an aspect discussed in section 3.

In this paper, we deal with usage of MNN concept for:

- Feature extraction of patterns
- Mapping the extracted features of the patterns
- Construction of the pattern recognition architecture

## 2: PATTERN RECOGNITION AND MEMORY MAPPING

We construct a software architecture which does feature extraction coupled with memory mapping for a "pattern recognizer". This architecture is hierarchical and is constructed out of several modules each of these modules is a "Mirroring Neural Network" [7]. The task of the latter is to take a set of data and then reduce (compress) the data in such a manner that information is not lost, the compressed data is then passed on to the next module. The additional purpose of the pattern recognizer is to perform automatic memory maps, by this we mean: given a set of data which is connected to another set of data, the pattern recognizer associates the former set with the latter. For example, suppose we are given a set of data which represents the digitized data (say a .wav file) of the spoken word "Alice Simmons" and another set of data which is a digitized image of the person, "Alice Simmons", then the pattern recognizer can be taught to associate the voiced data to her image. The pattern recognizer can then learn to associate any other speech signal containing the spoken word "Alice Simmons" to any other image of "Alice Simmons". The type of pattern recognizer described in this example can be utilized to construct an intelligent device which associates a spoken word to an image.

The main characteristics of the architecture are: (i) its input independence (ii) its hierarchical structure (iii) its modularity, (being built up of individual modules or "cells") and (iv) the attribute that each of these modules actually implements a single common algorithm (the MNN algorithm). Since the architecture has these 4 characteristics, it can be easily generalized and enlarged to solve a large variety of problems of increasing complexity and size. The single common algorithm that is used by the MNN which perform a compression (feature extraction) and a memory association of the data, details are below.

## 2.1. Overview of the proposed architecture

In order to design our architecture which has the above mentioned 4 characteristics, we have imitated some features of the neo-cortex [11] in the human brain, which seem to possess these characteristics. The human brain can encode the data of different sensory input patterns and not only recognize the patterns but is also able to effectively store these in very efficient memory maps and is capable of self-learning and decision making. Of course, our architecture can at best be said to be a crude copy of the complicated architecture of the neo-cortex whose details are yet unknown and are a subject of intense research among neuroscientists.

In our approach the hierarchical architecture is built up of modules connected in a tree-like structure each module is a MNN, the bottommost level is used to accept the sensory input data and extract those features that best capture the patterns which are to be recognized. At this level, each of the MNNs processes a category of sensory patterns. The extracted features of dissimilar sensory patterns from the lower level MNNs are used by the upper level MNNs so that the latter can function as associative memory maps. That is, in the upper level each MNN maps the extracted features of patterns belong to a group of sensory input to the features of the corresponding pattern group of a different sensory input. (For example, a sensory input representing the spoken word 'face" is associated with its corresponding visual sensory input i.e., "face image"). This process of feature extraction and the associated mapping can be repeated for each of the distinguishable sensory patterns. And we can say that our architecture performs a combination of learning and mapping thus simulating the processes of automatic learning, memorization and recognition as is done by the cortical system. An illustrated pictorial diagram (Figure 1) of the proposed architecture with explanation on sample implementation is given below.

Let us, for the sake of simplicity, consider only the two categories of sensory input patterns out of N categories (an example case, for the human brain the distinct categories of sensory patterns belong to Sound, Image, Odor, Taste, Sense/Touch) in our architecture, it will become apparent that there is no loss of generality in choosing only two categories. One of the two input patterns is a voice (sound) pattern and the other is an image pattern and they may be termed Category I and Category II respectively (this simplified situation corresponds to N=2, in Fig 1). Further, we assume that each of these categories contain 3 distinguishable pattern groups in it, say "Face", "Window" and "Garden" (this corresponds to k=3, in Fig. 1). Then the bottommost level MNNs i.e., MNN I and MNN II are trained to correspondingly extract the best low-dimensional features of voice patterns ('Face', 'Window' and 'Garden') and image patterns ('Face', 'Window' and 'Garden'). These extracted features are classified into the 3 distinguishable groups by Forgy's clustering algorithm [13] and resulting groups are given to the MNNs at the upper level. These upper level MNNs are then trained to map the input sound to the corresponding image from left to right at Level II. For example if the input data (in a .wav file) is the sound "window" it is automatically associated with the image of a "window". In this paper the first level of training is supervised in the sense that there is a training session wherein data is fed in pairs, one containing a spoken word and the other an image. For example, a spoken word "garden" (from a .wav file) is presented to the MNN I along with the image of a "garden" (from a .bmp file) to MNN II at level I. During the training many such pairs of data (speech-cum-image) belong to different groups (in our example, there are 3 groups) are presented, after that both the MNNs, MNN I and MNN II learn to classify the speech and images respectively. At level II these speech data and image data are associated by another set of MNNs - one MNN for each group.

After successful training, the pattern recognizing architecture will perform as follows: If a word (say a "window") is sent as input (to MNN I), then the network architecture will automatically associate this word with the image "window" and the .bmp file containing that image is displayed. Thus the architecture maps speech data to visual data.

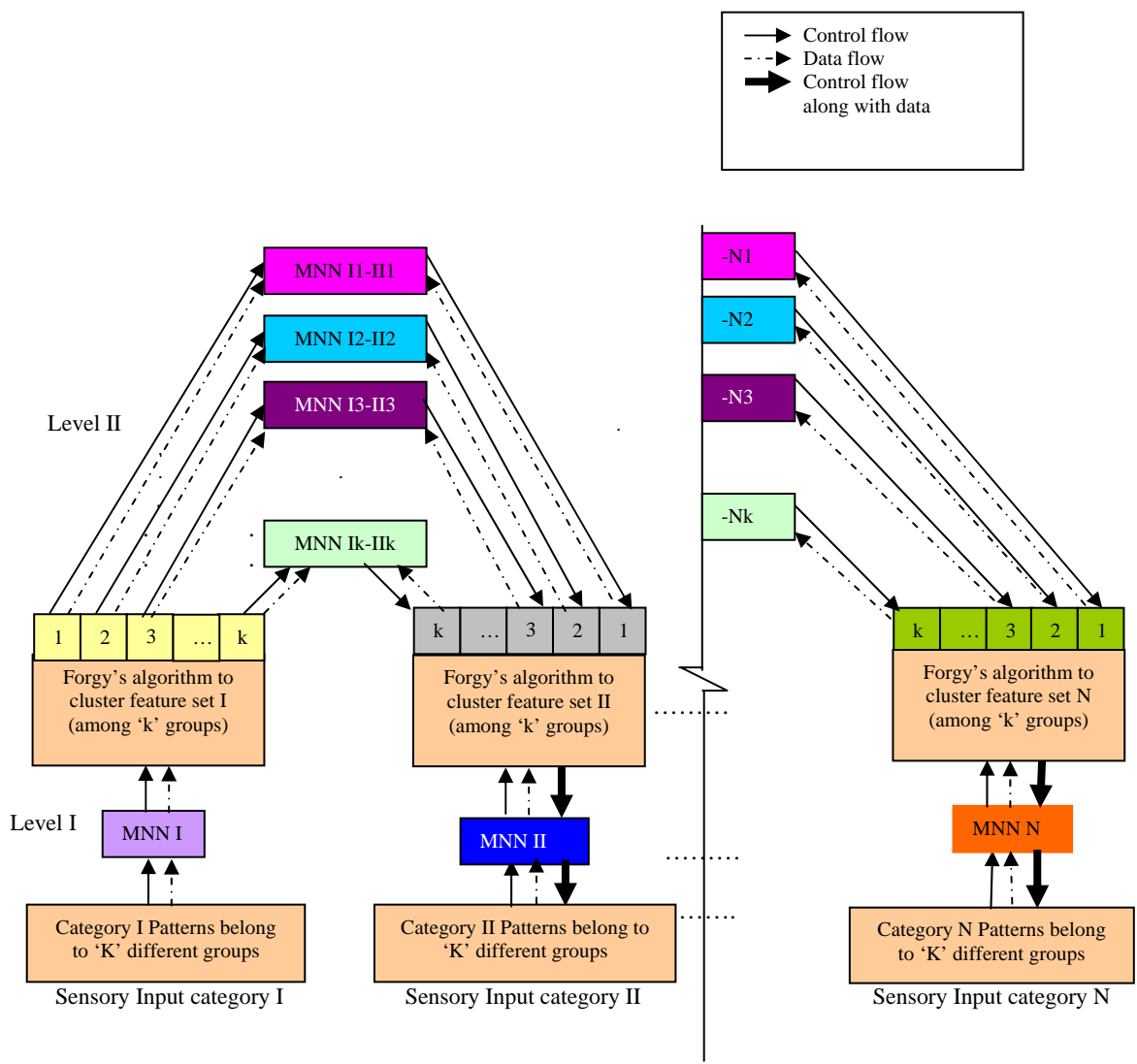

**Figure 1. Illustrated pictorial diagram of the proposed architecture**

### 2.2. The MNN concept

In this paragraph we briefly describe the functions of the Mirroring Neural Network (MNN), details are given in [7] and [8]. The MNN module is essentially a converging-diverging type of artificial neural network architecture. The MNN compresses the received input and gives an output. The idea is to train the MNN so that the output is as close as possible to the input. If this is successful for a particular class of patterns then the particular MNN has "learnt" the pattern, the output of the lowest dimensional hidden layer becomes the feature set characterizing the pattern. This feature set can then be used to distinguish it from other patterns and classify it. So we see an MNN does the following tasks: (i) compresses the input data, (ii) extracts a suitable feature set characterizing the input pattern and (iii) has the property to reconstruct the original data given the compressed data. It may be noted, one MNN can be used to recognize either one pattern or a particular pattern from a set of patterns. We use a MNN as a module of our pattern recognition architecture.

To summarize this section we can say that we implement the MNN's feature extraction (at level I) on different kinds of patterns i.e., voice samples besides image patterns. In addition to usual feature extraction, at the upper level, the MNN concept is to carry out a memory map by connecting two categories of sensory patterns which are associated with each other. And we construct the architecture for pattern recognition with the combination of lower and upper level MNNs which perform dimensional reduction and memory mapping respectively.

## 2.3. Demonstration of the proposed architecture

In this section, we describe the procedure followed for training the pattern recognition network and present the results on an actual example.

**2.3.1. Inputs to Level I.** The inputs to the architecture are pairs of voice samples and their corresponding images. The voice samples are the actual spoken words repeatedly uttered by a particular speaker. Actually we took 150 samples of the word "face"( i.e., the word "face" repeatedly uttered 150 times) similarly 150 samples of the word "window" and the 150 samples of the word "garden" are taken. Each of these words is paired with a corresponding image data and this data pair is fed to the architecture as Sensory input I (voice) and sensory input II (image).

The procedure to obtain the digitized voice samples is as follows: each voice sample is digitized at 2000 samples per second and re-sampled to 510 equally spaced points using Matlab [14]. These 510 values of the word samples are considered as representing the sensory input I (input vector of 510 dimensions to MNN I).

For inputting the data for the images the procedure is as follows: A total of 450 different images were considered in this study, all the images though different, fall into one of three classes viz., a face (faces are from the Yale face database [15]), a window or a garden. Each of these images is resized to a fixed size 17 X 30, containing 510 pixels. The grayscales (intensity levels of each pixel) whose range is from 0 to 255 are rescaled [16] so that they all lie between -1 to +1, these 510 intensity values constitute the input vectors for each sample image and are given as input vectors to MNN II.

**2.3.2. Classification at Level I.** During the training period when pairs of (voice, image) data are fed to level I, both MNN I and MNN II learn (i) to reduce the input data and then (ii) to independently classify their respective inputs.

The data reduction is done from 510 to 20 units by using a 3 layer MNN architecture, 510-20-510 processing elements. The reduced inputs are then used to classify the inputs. For example the 20 features extracted by the MNN I classifies its input words, using Forgy's algorithm into 3 groups (k=3 in figure) for the 3 types of words "face", "window" and "garden". Similarly, and MNN II reduces its image data to 20 features using again a 510-20-510 architecture and classifies the inputs into 3 groups one for a "face image", another for "window image" and the third for the "garden image". See[7] and [8] for more details on data reduction by MNN and its use of Forgy's algorithm.

**2.3.3. Training of MNN's in Level II.** In the following example the Level I MNNs are first trained and their inputs classified and then the level II MNN's are trained for associating words with the corresponding images.

But, it is to be mentioned here that it is quite possible to envisage both the levels to be trained simultaneously, under the assumption of simultaneity by using temporal information [10] and [11]. If we assume that the sensory input I is related to Sensory input II, this will happen if the word face is presented simultaneously with the image of a face, then MNN I and MNN II can then classify their inputs and put them in the same group (say group 1 for face), simultaneously MNN I1-II1 in Level II will be trained such that the reduced input given to MNN I1-II1 (from the MNN I in Level I) is mapped (matched) to the reduced input of MNN II at level I. If this is done properly, then the architecture will train itself to associate the word face with the image of a face and the word window with the image of a window etc.,.

The purpose of Level II is to associate each group of input from Sensory I to its appropriate group from Sensory II. That is, if the word "face" is invoked as input, then the architecture should associate a "face image" to this input. Since there are 3 groups (k=3) in our data, there will be 3 MNNs in Level II each of these have to be trained with the appropriate inputs. We have chosen a 20-20-20 architecture for the 3 MNNs in Level II.

The procedure to train these level II MNNs is as follows: the reduced input from MNN I (in this case it is a 20 dimensional vector), becomes the input to the appropriate MNN in Level II. Example, if the input (garden word, garden image) is fed as data to level I MNNs then the 20 feature vector (of garden word) from MNN I is given as input to MNN I3-II3 in Level II so that its output is equal to the 20 dimensional feature vector of the garden-image, obtained by data reduction using MNN II in Level I.

After the training of the Level II MNNs, the whole architecture can be considered as trained. The functioning of this trained architecture can be now tested. Simply by inputting a voice sample and finding out if the architecture maps this voice input to its corresponding image and give it as an output. The flow chart of the architecture is shown in Figure 2.

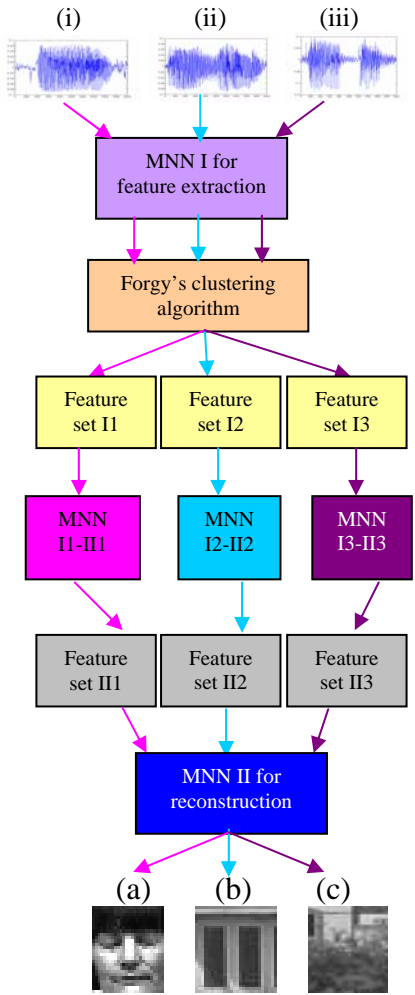

**Figure 2. Pictorial representation of pattern recognition and memory mapping exemplified on typical samples from each group: (i) word 'face' mapped to image 'face' (a); (ii) word 'window' mapped to image 'widow' (b); (iii) word 'garden' mapped to image 'garden' (c).**

**2.3.4 Results.** The experiment is conducted by using a data set of 450 (word, image) pairs, for training purposes 300 pairs are used and then the remaining 150 pairs are used for testing (evaluation). It is emphasized here that the pairs that were used for testing the architecture were new pairs of data and were NOT used while training the architecture. Table 1 summarizes our results.

At Level I, the efficiencies of the MNN I and MNN II in classifying the voice and image patterns into their appropriate group is found to be 91.6% and 95.3% respectively (using only the reduced input vector of 20 dimensions).

The overall efficiency of recognition, that is, the rate of correct prediction of a voice input to its appropriate image output is found to be 91.6%.

**Table 1: Pattern recognition and memory mapping using MNN**

| Input to the system | Output from the system | Success rate of MNN I | Success rate of MNN II | Over-all efficiency |
|---|---|---|---|---|
| Voice samples | Image samples | 91.6% | 95.3% | 91.6% |

### 3: CONCLUSIONS AND FUTURE WORK

We have demonstrated the successful functioning of an unsupervised learning algorithm which has the following features: (i) It is hierarchical and modular (ii) each module runs on a common algorithm, (iii) capable of automatic data reduction and feature extraction and (iv) provides an efficient associative memory map. Because of these features it is capable of being enlarged and used for more complex learning tasks. For example, its ability to associate a voice pattern with an image pattern makes it a good candidate for devising learning machines which can associate memories of two simultaneous events e.g., the image of a train moving with the whistle blowing, the sight of fire with its heat. This kind of learning machine could provide a method of associating memories of two different events separated temporally and thus learn to recognize cause and effect. It is hoped that architecture is flexible enough to be deployed, in the future, for even more complex pattern recognition tasks as those performed by neo-cortex in the human brain.